# Deep Recurrent Neural Networks for ECG Signal Denoising

Karol Antczak [a,1]

[a] *Institute of Computer and Information Systems, Military University of Technology, Kaliskiego 2, 00-908 Warsaw, Poland*
[1] Corresponding author. *E-mail address*: karol.antczak@wat.edu.pl

ABSTRACT

**Background and Objective:** Electrocardiographic signal is a subject to multiple noises, caused by various factors. It is therefore a standard practice to denoise such signal before further analysis. With advances of new branch of machine learning, called deep learning, new methods are available that promises state-of-the-art performance for this task.
**Methods:** We present a novel approach to denoise electrocardiographic signals with deep recurrent denoising neural networks. We utilize a transfer learning technique by pretraining the network using synthetic data, generated by a dynamic ECG model, and fine-tuning it with a real data. We also investigate the impact of the synthetic training data on the network performance on real signals.
**Results:** The proposed method was tested on a real dataset with varying amount of noise. The results indicate that four-layer deep recurrent neural network can outperform reference methods for heavily noised signal. Moreover, networks pretrained with synthetic data seem to have better results than network trained with real data only.
**Conclusions:** We show that it is possible to create state-of-the art denoising neural network that, pretrained on artificial data, can perform exceptionally well on real ECG signals after fine-tuning.

## 1. Introduction

Electrocardiography (ECG) is a diagnostic process that records the electrical activity of the heart over time. The measurement is done by electrodes attached to the patient's body and a medical device called electrocardiograph. Nowadays there are typically 10 electrodes used, placed on the chest and limbs. A result of the procedure is the electrocardiogram - a graphical visualization (waveform) of changes in electrical potential of the heart. For 10-electrode measurement, electrocardiogram contains 12 waveforms, one for a specific combination of electrodes, called "leads". Each lead represents the electrical activity of the heart from a specific angle.

ECG waveform is a quasi-cyclical time series. Its values represent electrical potential measured for the specific lead over some period of time, usually ~10 seconds. Each quasi-cycle on the waveform represents one cardiac cycle, therefore it has a typical frequency between 0.67 to 5 Hz. In addition to the main cardiac cycle, the signal also contains other non-noise components that range from 0.67 to 500 Hz.

The main cycle (Figure 1) has characteristic components named P, Q, R, S and T, corresponding to specific events during the cardiac cycle. Three of them: Q, R and S create so-called QRS complex – a main "spike" in the cycle, corresponding to a sudden depolarization of ventricles. Visual analysis of properties of QRS complex and other points is usually performed manually by diagnostician and allows to identify abnormalities and diseases associated with them.

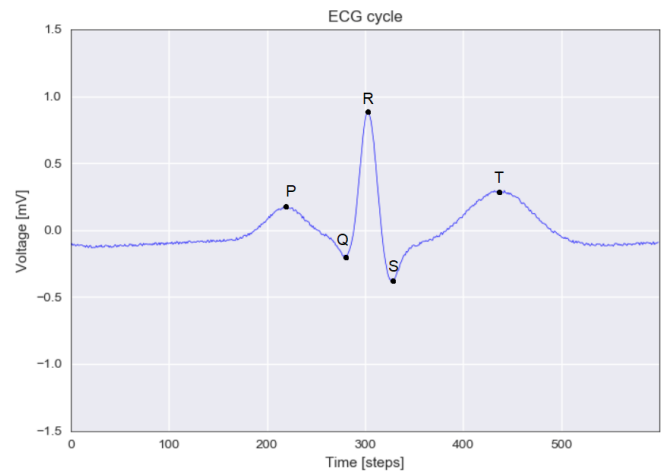

**Figure 1. A single ECG cycle.**

ECG signal is a subject to various noises with characteristic frequency spectrums. [1] provides a list of several types of noise in ECG, depending on the source. They are:

- Power line interference - results in random component at 60 or 50 Hz, depending power supply frequency
- Electrode contact noise - caused by improper contact of between the body and electrodes with ~1Hz frequency
- Motion artifacts - produced by patient's movements which affect electrode-skin impedance, resulting in 100-500 ms long distortions.
- Muscle contractions - muscle activity produces noise with 10% of regular peak-to-peak ECG amplitude and frequency up to 10 kHz. The duration is typically around 50 ms.
- Baseline wander - caused by a respiratory activity, having 0-0.5 Hz frequency. The amplitude of this noise is ~15% of overall ECG amplitude.

A goal of ECG signal denoising is to remove these noises while keeping as much of the signal as possible. Since frequencies of the signal and the noise overlap, this is a challenging task. This paper introduces a new way to denoise the signal, utilizing deep recurrent neural networks (DRNN). The network is trained using two datasets: a synthetic one and a real data. We study how using synthetic dataset affect the network performance.

The structure of this paper is as follows: in section 2 we review existing approaches to denoising ECG signal and related applications utilizing neural networks. In section 3 we discuss a dynamic model for generation of synthetic ECG signals. In section 4 an idea of deep recurrent denoising network for ECG processing is presented. Section 5 contains test results. In section 6 we present conclusions.

*1.1. Related works*

Denoising ECG data is a known problem with relatively long history; a number of techniques therefore exists. Common approaches are listed in [2]. They include finite and infinite impulse response filters, wavelet-based methods, filtered reside technique and empirical mode decomposition.

In recent years, one can observe an appearance of novel approaches to signal filtering, utilizing machine learning methods and neural networks. For example, Moein [3] investigated multi-layer perceptron networks for ECG noise removal. For training, a relatively small dataset of 100 signal samples was used. The expected outputs were produced by denoising input signals using Kalman filter. The network was able to achieve error rate less than 0.5 for all of them. However, it is worth noting that due to the nature of the dataset, the network learned in fact to simulate the Kalman filtering. Therefore, by training the network this way, one cannot achieve better performance than the Kalman filtering itself.

Another approach to neural network-based noise reduction is described in [4]. It utilizes both neural networks and wavelet transform, in a form of Wavelet Neural Networks (WNN). Such networks are a special kind of three-layer feedforward neural networks, employing a set of wavelets as activation functions. The network training is a two-phase process; first, using 400 iterations of specialized algorithm – Adaptive Diversity Learning Particle Swarm Optimization (ALDPSO) – that performs global search in the population of 20 candidate networks. The second phase are 1600 iterations of gradient descent of the best-performing network from previous phase. The training and test data are real signals from PhysioBank database; however, there is no information about amount of the data used. The network input is then noised using with a white noise of signal-to-noise ratio (SNR) 17.7 dB, and expecting output is a clean, unprocessed signal. The trained network is able to filter a signal to have approximately 21.1 dB SNR, which is 4.1. dB of improvement.

To the best of our knowledge, there are not recorded attempts usage of DRNNs for the specific purpose of denoising of ECG signal. Nonetheless, in general, recurrent neural networks are used quite often for signal processing purposes, including signal denoising. As of September 2018. Google Scholar website yields 1820 results for the query "deep RNN for denoising". Known applications of these networks include acoustic signals [5] or videos [6]. A relatively popular approach recently is to combine LSTM-based RNNs with autoencoder-like training, resulting in deep recurrent denoising networks [7] [8] [9]. The recurrent network with this architecture is trained to recreate the noised input signal.

## 2. Methods

*2.1. Deep recurrent neural network*

We propose to use deep recurrent denoising neural networks (DRDNN) for denoising of ECG signal. They are a kind of deep recurrent neural networks (DRNN), and, as such, have two distinct features. The first one is that they consist of multiple (> 2) layers stacked together

- this approach is also known as a deep learning. Such deep architectures, while requiring more computational power than typical "shallow" neural networks, were proven to be highly effective in various application, due to their ability to learn hierarchical representation of the data - initial layers learn "simple" features, while next layers learn more complicated concepts. Interestingly enough, it is still not known in general why deep neural networks are more effective than shallow ones. Several hypotheses were proposed [10] [11] [12], but the research is still ongoing.

The second distinct feature of DRNNs (and all recurrent neural networks in general) is their ability to preserve its internal state over time. It is usually obtained by introducing recurrent connections in the network, that return the previous output of the neuron to itself and/or other units in the same layer. This makes them a common choice for machine learning tasks involving processing or prediction of sequences and time series [13].

A popular architecture of deep recurrent networks involves a specific kind of building blocks called Long Short-Term Memory (LSTM) units [14]. As the name suggests, they remember its internal state for either long or short period of time. A typical LSTM is composed of four components: a memory cell, and three gates: input, output and forget. Each gate is connected with other through its input and output; several connections are recurrent. Processing of the LSTM unit updates its internal state $c_t$ (memory cell) and, simultaneously, produces the output vector $h_t$, according to the set of equations:

$$f_t = \sigma_g(W_f x_t + U_f h_{t-1} + b_f)$$
$$i_t = \sigma_g(W_i x_t + U_i h_{t-1} + b_i)$$
$$o_t = \sigma_g(W_o x_t + U_o h_{t-1} + b_o)$$
$$c_t = \sigma_c(W_c x_t + U_c h_{t-1} + b_c) \circ i_t + f_t \circ c_{t-1}$$
$$h_t = o_t \circ \sigma_h(c_t)$$

$x_t$ is the input vector, while $f_t, i_t$ and $o_t$ are activation vectors of input, forget and output gates, respectively. $W$ denotes weight matrices of respective gates. The operator ∘ denotes the Hadamard product. $\sigma_g$, $\sigma_c$ and $\sigma_h$ are activation functions. In a typical implementation, $\sigma_g$ is the sigmoid function while $\sigma_c$, $\sigma_h$ are hyperbolic tangent functions.

LSTM units can be connected into a larger structure in two ways. The first one is connecting output of cell memory and hidden gate of one cell to the input and output of the forget gate, resulting in a single LSTM layer. Such layer can be then stacked by connecting inputs of one layer to output of the next layer. This allows to build multilayer, deep LSTM network.

The input signal to LSTM network is applied one sample at a time. It is then propagated through each layer and result a single output in each iteration. Therefore, the output signal of N length can be obtained by applying the input of the same length. This is perfectly fine for denoising purposes such as ours, however it is worth nothing that for classification of prediction tasks, the signal needs to be "flattened". Therefore, many DRNN-based architectures contain also some number of feedforward layers.

LSTM networks can be trained using backpropagation through time (BPPT) and its variants [15]. This algorithm works similar to classical backpropagation but operates on the unfolded structure of the recurrent network. This works because every recurrent network can be "unfolded" into equivalent feedforward network. The loss function is calculated as an average cost for all time steps in the sequence.

In this paper, we use the deep recurrent denoising neural network, which is a specific hybrid of DRNN and a denoising autoencoder. The denoising autoencoder is a neural network trained to recreate noised input data, with input layer of the same width as the output layer. DRDNN are therefore DRNN with input layer of the same shape as the output layer, trained by applying noised signal to the input and expected to produce its denoised equivalent at the output. General architecture of DDRNNs for ECG denoising proposed in this paper is presented in Figure 2. The ECG signal can be represented as a one-dimensional vector $x$. It is applied to the network and outputted by it, element-by element at each time step $t$, resulting in the output vector $y$. Therefore, the network has both input and output of width 1. The signal is first processed by the recurrent layer consisting of LSTM units. Next, the signal is processed by a certain number of dense layers with rectified linear activations. The last layer is a linear one, which simply sums up output from the previous layer.

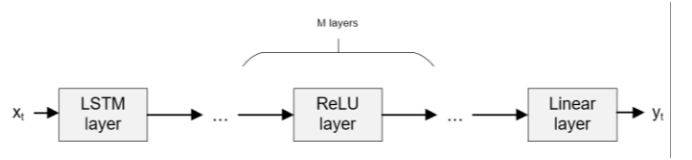

**Figure 2. Architecture of deep denoising recurrent neural network.**

## 2.2. Data generation and preprocessing

For generation of synthetic training data, we used a dynamic model described in [16]. It allows to generate a realistic ECG signal basing on statistical properties of the signal, like mean and deviation of heart rate or low/high frequency power ratio. Additionally, the model incorporates a set of morphological parameters of P, Q, R, S and T events that can be specified. Finally, it is possible to define measurement parameters of generated signal, like signal sampling frequency and measurement noise.

The model is described by a set of three differential equations:

$$\dot{x} = \alpha x - \omega y$$
$$\dot{y} = \alpha x + \omega y$$
$$\dot{z} = -\sum_{i \epsilon \{P,Q,R,S,T\}} a_i \Delta\theta_i \exp\left(-\frac{\Delta\theta_i^2}{\Delta b_i^2}\right) - (z - z_0)$$

where

$$\alpha = 1 - \sqrt{x^2 + y^2}$$
$$\Delta\theta_i = (\theta - \theta_i) \bmod 2\pi$$
$$\theta = \mathrm{atan2}(y, x)$$

Above equations describe a trajectory of a point in 3D space with coordinates $(x, y, z)$. The trajectory is cyclical, revolving around a limiting circle of unit length. This reflects the quasi-periodicity of the signal.

The baseline wander of the ECG incorporated into a model by defining $z_0$ as a periodic function of time:

$$z_0(t) = A \sin(2\pi f_2 t)$$

where $A$ is the signal amplitude (in mV) and $f_2$ is the respiratory frequency.

The model defines also a power spectrum $S(f)$ of the signal. It is a sum of two Gaussian distributions:

$$S(f) = \frac{\sigma_1^2}{\sqrt{2\pi c_1^2}} \exp\left(\frac{(f-f_1)^2}{2c_1^2}\right) + \frac{\sigma_2^2}{\sqrt{2\pi c_2^2}} \exp\left(\frac{(f-f_2)^2}{2c_2^2}\right)$$

where $f_1, f_2$ are means, $c_1, c_2$ are standard deviations and $\sigma_1^2, \sigma_2^2$ are powers in low- and high- frequency bands, respectively.

To use a model in practice, several parameters needs to be specified. The main goal here is to obtain a signal as similar to the real data as possible. Morphological parameters $(a_i, b_i, \theta_i)$ were set for each of P, Q, R, S and T events, with the same values as in [16]. Frequency and spectral parameters were also mostly the same, with the exception of heart rate standard deviation - value of 5 was used. Moreover, the sampling rate used was 512. The signal had additive uniform (white) noise with amplitude 0.01. It was treated as a "base" noise level – a different from the one being added later for denoising task. A sample of generated synthetic signal can be seen in figure below.

Real dataset used in this research came from Physionet PTB diagnostic database [17]. It contains 549 records from 290 subjects. Signal in database are available for 15 leads. For our purposes, we decided to use aVL lead only.

The signals, be it artificial or real, were preprocessed by normalizing them to have a zero mean. Next, they were divided into samples 600 datapoints each. Such number of points corresponds to approximately two ECG cycles. Such preprocessed datasets were used as expected outputs during training and testing of networks. The inputs signals for networks were produced by adding a white noise to the reference signal. All metrics were then calculated on the normalized data.

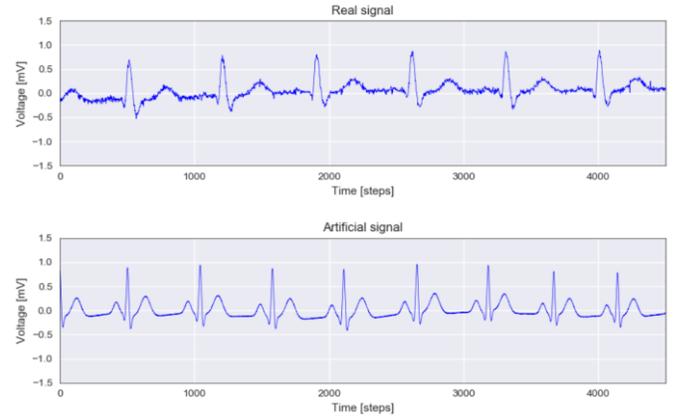

Figure 3. Examples of real (top) and artificial (bottom) ECG signal.

## 2.3. Reference methods

For comparison purposes, we used two reference methods: bandpass filter and Undecimated Wavelet Transform (UWT) methods. Each of them has its advantages and disadvantages. Bandpass filter is a combination of low- and high- pass filters that cuts out all frequency components not fitting into a certain range. The frequency range used depends on the purpose of the signal capturing and patient characteristics. Recommendations given by American Heart Association for diagnostic electrocardiography of adults, adolescents and children suggest using 0.05-150 Hz range [18]. This range is used in our implementation as well.

A more sophisticated approach to ECG denoising is based on Undecimated Wavelet Transform (UWT), described by Hernández and Olvera [19]. The main principle of the method is signal decomposition using stationary wavelet transform:

$$\omega_v(\tau) = \frac{1}{\sqrt{v}} \int_{-\infty}^{+\infty} s(t) \psi * \left(\frac{t-\tau}{v}\right) dt$$
$$v = 2^k, k \in \mathbb{Z},$$

where $\omega_v$ are the UWT coefficients, $v$ is the scale coefficient, $\tau$ is the shift coefficient and $\psi *$ is the complex conjugation of the mother wavelet. As a mother wavelet, Daubechies D6 wavelet was used, due to its similarity to the ECG signal.

Signal decomposition in UWT is an iterative process, with each iteration producing the approximation of the signal and detail coefficients for given level $k$ of decomposition (see Figure 4). After the decomposition, the signal is filtered by removing coefficients below the threshold $T$. The signal is then reconstructed from coefficients by performing inverse wavelet transform.

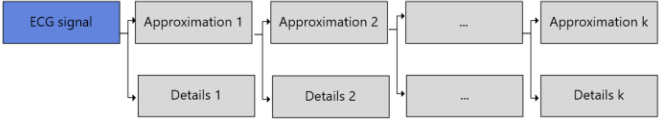

**Figure 4.** ECG signal decomposition with Undecimated Wavelet Transform.

In our implementation, two customizations of the original method were used in order to adjust the method to our needs. First, a five-level decomposition was used, as it was empirically determined to yield the best results. Second, the universal threshold proposed by Donoho and Johnson [20] was used, defined as:

$$T = \sigma\sqrt{2\log_e N}$$

where $\sigma$ is the median absolute deviation of coefficients and $N$ is the number of data points.

### 2.4. Performance measures

Two metrics were used for model performance measurement: mean squared error and signal-to-noise ratio:

$$MSE = \frac{1}{n}\sum_{i=1}^{n}(s_i - x_i)^2$$

$$SNR = 10\log_{10}\frac{\sum_{i=1}^{n} x_i^2}{\sum_{i=1}^{n}(s_i - x_i)^2}$$

Mean squared error is used during pretraining and fine-tuning phase, since it serves as a loss function for weight update. Signal to noise ratio, on the other hand, was used for comparison of various denoising methods.

## 3. Results

### 3.1. Network architectures comparison

First, we evaluated networks of various sizes to find the one that will have the best performance while not being too big to train effectively. Two parameters were analyzed here: "width" (number of neurons/units per layer) and "depth" (number of ReLU layers) of the network. We trained networks with number of ReLU layers from 0 to 9 and having 16, 32 and 64 neuron per layer. Each network was trained on the 2000 samples and validated on 2000 samples, both of them being artificial ones. Each sample consisted of 600 data points. Training algorithm used was stochastic gradient descent with adaptive momentum (Adam). 64 samples per-batch were used. The loss metric was mean squared error. Training length was 50 epochs in each case.

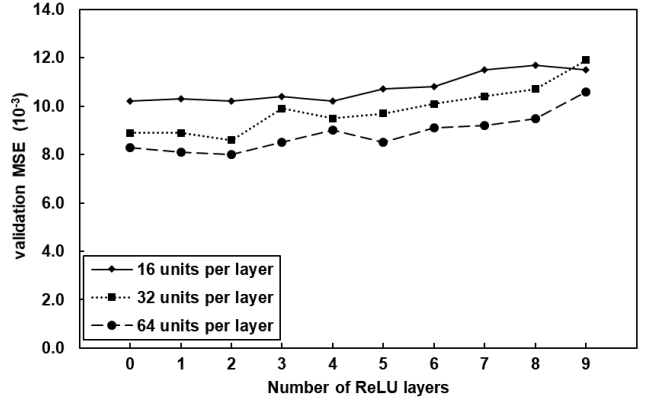

**Figure 5.** Performance of DRNNs of various architectures.

Results can be seen in Figure 5. One can observe that increasing a number of neurons per layers improved network performance in general. However, this, increased the training duration as well - networks with 64 neurons per layer were trained about 50% longer than their counterparts with 16 neurons per layer. In terms of number of layers, the best results were obtained for 2 ReLU layers. Having more than 5 layers worsened performance regardless of number of neurons per layers. The best network overall was the one with LSTM layer with 64 units followed by 2 ReLU layers, 64 units each and a single linear layer.

### 3.2. Effect of pretraining

After selecting the best network architecture, we analyzed how pretraining the network on the artificial signal affects its performance after fine-tuning on real data. To achieve this, we prepared a set of neural networks pretrained with 0, 10 and 50 epochs, respectively and then fine-tuned each of them with real data. Pre-training data was artificial dataset used in previous test, with 64 samples per patch. Fine-tuning dataset consisted of 5000 real training samples and 2000 reals validation samples.

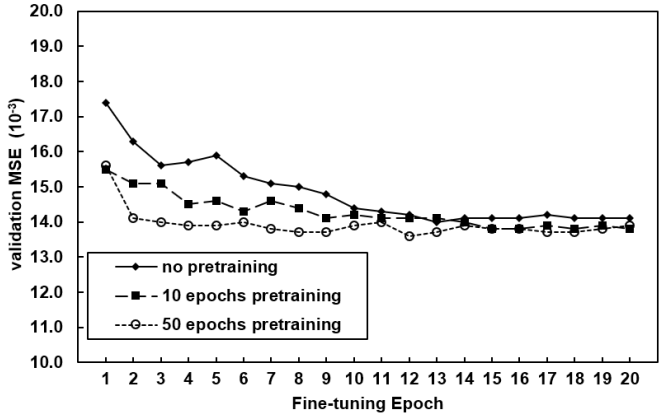

**Figure 6. Effect of pretraining on network performance.**

Results, seen in the figure above, indicate that pretraining the network on artificial data indeed helps in fine-tuning the model. Pretrained networks were able to converge significantly faster compared to non-pretrained one. The biggest difference was seen in initial epochs. After ~10 epochs, networks losses converged to similar values. Nonetheless, at the end of the fine-tuning, pretrained networks were able to achieve lower values, obtaining 0.0139 error for network pretrained with 50 epochs, while non-pretrained network obtained 0.141 error.

*3.1. Comparison with reference methods*

With the fine-tuned network, we compared it with reference methods – wavelet filter and bandpass filter - for denoising the signal with varying signal-to-noise ratio. The test dataset consisted of 5000 real samples with signal-to-noise (S/N) ratios from -20 dB to 0 dB. We then measured S/N of the output signal. Results can be seen in Figure 7, while visual examples of denoising selected samples are presented in Figure 8.

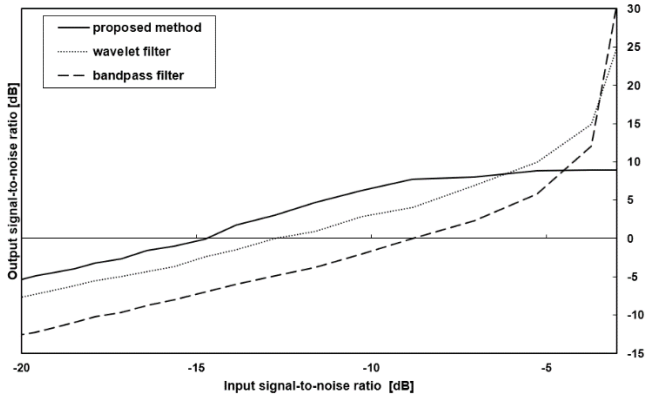

**Figure 7. Performance of denoising methods for various signal-to-noise ratios.**

For signal with relatively small noise (>-3dB) the best results were achieved by a simple bandpass filter. However, this method seemed to be the most sensitive to the noise and observed a significant loss of performance for higher noise levels. The best denoising methods for signals with noise level from -3 dB to -7 dB was the wavelet filter. For noise levels < -7 dB, proposed deep neural network started to outperform reference methods. An interesting property of the neural network was its stability for different noise levels. It obtained almost the same results for noise levels from -9 dB to 0 dB.

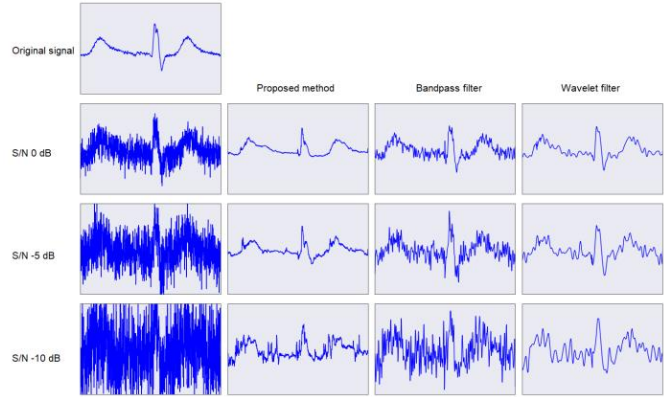

**Figure 8. Visual examples of ECG denoising results.**

## 4. Conclusions

Obtained results show that DRDNN can be used for effective denoising of ECG signals, obtaining 7.71 dB signal-to noise-ratio from the input signal with -8.82 dB S/N, outperforming reference methods. It is therefore yet another area where deep networks show their superiority over shallow architectures. However, increasing number of layers causes a risk of network overfitting. It could be overcome using regularization techniques for deep learning such as L1/L2 penalties, drop-out or drop-connect. It is possible that tuning up hyperparameters would result in even better performance than obtained in this paper.

Another important conclusion comes from the analysis of influence of synthetic training data on network performance. The outcome suggests that networks trained with artificial data have better performance than networks trained with real signals only. This can be partially explained by the interpreting the training process by means of "transfer learning" framework [21]. It is a popular deep learning technique that allows to train the network using training data with different domain, distribution and task than the target data. The network can be than applied to the target task with relatively small amount of fine-tuning. Transfer learning is explained by the analogy of human learning process: people can use previously gained knowledge to solve problems faster, even if such knowledge was acquired for different domain. This explains that network trained with synthetic ECG data were able to denoise real signal as well.

Still, the transfer learning hypothesis in its original form does not explain why pretraining with artificial data was more beneficial than using real data only, even though lack of pretraining was partially compensated in later epochs of fine-tuning. We propose to explain this by using another analogy to the human cognitive process. It is natural for people to learn by observing simpler examples first and more complicated ones later. This allows to gain knowledge incrementally, by grasping the "essence" of the knowledge first and then fine-tuning it with more complex examples. Learning from the complex examples usually yields rather mediocre results. In terms of network training, synthetic ECG data was based on some mathematical model. Models are, by definition, a simpler view of something more complicated. In our case, ECG signal model was a rather simple one, not including many bio- & electro- physical phenomena; moreover, it assumed a naive model of noise. However, due to its simplicity, it was easier to learn than real data, as indicated by training results. Therefore, network trained on synthetic data was able to faster learn features of the ECG signal and utilize this knowledge for learning from real data. This implies that using artificial training data (by means of transfer learning) is a promising approach not only in situations of data shortage (which is often the case in a medical field) but also to improve the quality of the network with the a priori knowledge included in the mathematical model of the data.

## Conflict of interest

The author does not have financial and personal relationships with other people or organizations that could inappropriately influence (bias) their work.

## Acknowledgements

The author would like to thank Piotr Bączyk for his suggestions regarding reference filtering methods.